\documentclass[]{article}

\usepackage[T1]{fontenc}
\usepackage{amsmath,amssymb}
\usepackage{graphicx}

\usepackage{hyperref}
\usepackage{color}
\usepackage{footmisc}

\usepackage{algorithm}
\usepackage{algpseudocode}

\usepackage[backend=biber,style=numeric]{biblatex}
\begin{document}
\title{Toward Design of \\ Synthetic Active Inference Agents \\ by Mere Mortals}

\author{Bert de Vries \\ 
Eindhoven University of Technology \\ Eindhoven, the Netherlands \\
\texttt{bert.de.vries@tue.nl}} 

\maketitle   

\begin{abstract}
The theoretical properties of active inference agents are impressive, but how do we realize effective agents in working hardware and software on edge devices? This is an interesting problem because the computational load for policy exploration explodes exponentially, while the computational resources are very limited for edge devices. In this paper, we discuss the necessary features for a software toolbox that supports a competent non-expert engineer to develop working active inference agents. We introduce a toolbox-in-progress that aims to accelerate the democratization of active inference agents in a similar way as TensorFlow propelled applications of deep learning technology.

\end{abstract}

\section{Introduction}\label{sec:intro}

This position paper aims to complement a recent white paper on designing future intelligent ecosystems where autonomous Active InFerence (AIF) agents learn purposeful behavior through situated interactions with other AIF agents \cite{friston_designing_2022}. The white paper states that these agents ``... can be realized via (variational) message passing or belief propagation on a factor graph'' \cite[][abstract]{friston_designing_2022}. Here, we discuss the computational requirements for a factor graph software toolbox that supports this vision. Noting that the steep rise of commercialization opportunities for deep learning systems was greatly facilitated by the availability of professional-level toolboxes such as TensorFlow and successors, we claim that a high-quality AIF software toolbox is needed to realize the proposition in \cite{friston_designing_2022}. Therefore, in this paper, we ask the question: what properties should a factor graph toolbox possess that enable a competent engineer to develop relevant AIF agents? The question is important since the number of applications for autonomous AIF agents is expected to vastly outgrow the number of world-class experts in AIF and robotics. 

As an illustrating example, consider an engineer (Sarah) who needs to design a quad-legged robot that is tasked to enter a building and switch off a valve. We assume that Sarah is a competent engineer with an MS degree and a few years of experience in coding and control systems. She has some knowledge of probabilistic modeling but is not a top expert in those fields. 


In order to relieve Sarah from designing every detail of the robot, we expect that the robot possesses some ``intelligent'' adaptation capabilities. Firstly, the robot should be able to define sub-tasks and solve these tasks autonomously. Secondly, since we do not know a-priori the inside terrain of the building, the robot should be capable of adapting its walking and other locomotive skills under situated conditions. Thirdly, we expect that the robot performs robustly, in real-time, and cleverly manages the consumption of its computational resources. 

 All these robot properties should be supported seamlessly by Sarah's AIF software toolbox. For instance, she should not need to know the specifics of how to implement robustness in her algorithms or how many time steps the robot needs to look ahead in any  given situation for effective planning purposes. We want a toolbox that enables competent engineers to develop effective AIF agents, not a toolbox for a select group of world-class machine learning experts. We do expect that Sarah is capable of describing her beliefs about desired robot behavior through the high-level specification of a probabilistic (world or generative) model or, at least, the prior preferences or constraints that underwrite behavior.

After reviewing some motivating agent properties that follow immediately from committing to free energy minimization (section~\ref{sec:FEP-and-AIF}), we proceed to discuss why message passing in a factor graph is the befitting framework for implementing AIF agents (section ~\ref{sec:why-MP}). More specifically, we argue that a reactive programming-based implementation of message passing will be the standard in professional-level AIF tools (section~\ref{sec:Reactive-vs-Procedural}). In comparison to the usual procedural coding style, reactive message passing leads to increased robustness (section~\ref{sec:robustness}), lower power consumption (section~\ref{sec:low-power-consumption}), hard real-time processing (section~\ref{sec:real-time-processing}), and support for continual model structure adaptation (section~\ref{sec:structural-adaptation}). In section~\ref{sec:rxinfer} we introduce \texttt{RxInfer}, a toolbox-in-progress for developing AIF agents that robustly minimize free energy in real-time by reactive message passing.

\section{The Free Energy Principle and Active Inference}\label{sec:FEP-and-AIF}

\subsection{FEP for synthetic AIF agents}

The Free Energy Principle (FEP) describes  self-organizing behavior in persistent natural agents (such as a brain) as the minimization of an information-theoretic functional that is known as the variational Free Energy (FE).\footnote{For reference, we use the following abbreviations in this paper: Active Inference (AIF), Constrained Bethe Free Energy (CBFE), Expected Free Energy (EFE), (variational) Free Energy (FE), Free Energy Principle (FEP), Free Energy Minimization (FEM), Message Passing (MP), Reactive Message Passing (RMP).} Essentially, the FEP is a commitment to describing adaptive behavior by Hamilton's Principle of Least Action \cite{ lanczos_variational_1986}. The process of executing FE minimization in an agent that interacts with its environment through both active and sensory states is called \emph{Active Inference} (AIF). Crucially, the FEP claims that, in natural agents, FE minimization is \emph{all that is going on}. While engineering fields such as signal processing, control, and machine learning are considered different disciplines, in nature these fields all relate to the same computational mechanism, namely FE minimization. 

For an engineer, this is good news. If we wish to design a synthetic AIF agent that learns purposeful behavior solely through self-directed environmental interactions, we can focus on two tasks:
\begin{enumerate}
    \item Specification of the agent's model and inference constraints. This is equivalent to the specification of a (constrained) FE functional. 
    \item A recipe to continually minimize the FE in that model under situated conditions, driven by environmental interactions.
\end{enumerate}

We are interested in the development of an engineering toolbox to support these two tasks.

\subsection{FEM for simultaneous refinement of problem representation and solution proposal}\label{sec:one-solutiona-approach}

An important quality of the robot will be to define tasks for itself and solve these tasks autonomously. Here, we shortly discuss how the FEP supports this objective.  

Consider a generative model $p(x,s,u)$, where $x$ are observed sensory inputs, $u$ are latent control signals and $s$ are latent internal states. For notational ease, we collect the latent variables by $z = \{s,u\}$. The variational FE for model $p(x,z)$ and variational posterior $q(z)$ is then given by 
\begin{subequations}
\begin{align}
    F[q,p] 
    &= \underbrace{-\log p(x)}_{\text{surprise}} + \underbrace{\sum_z q(z) \log \frac{q(z)}{p(z|x)}}_{\text{bound}}  \label{eq:bound-evidence}\\
    &= \underbrace{\sum_z q(z) \log \frac{q(z)}{p(z)}}_{\text{complexity}} - \underbrace{\sum_z q(z) \log p(x|z)}_{\text{accuracy}} \label{eq:complexity-accuracy}\,.
\end{align}
\end{subequations}
The FE functional in \eqref{eq:bound-evidence} can be interpreted as the sum of surprise (negative log-evidence) and a non-negative bound that is the Kullback-Leibler divergence between the variational and the optimal (Bayesian) posterior. The first term, surprise, can be interpreted as a performance score for the problem representation in the model. This term is completely independent of any inference performance issues. The second term (the bound) scores how well actual solutions are inferred, relative to optimal (Bayesian) inference solutions. In other words, the FE functional is a universal cost function that can be interpreted as the sum of problem representation and solution proposal costs. FE minimization leads toward improving both the problem representation and solving the problem through inference over latent variables. In particular, FE minimization over a particular model structure $p$ should lead to nested sub-models that reflect the causal structure of the sensory data. Sub-tasks are solved by FE minimization in these sub-models. Hence, both creation of subtasks and solving these subtasks are driven solely by FE minimization. 


In conclusion, a high-end toolbox should be capable to minimize FE both over (beliefs over) latent variables through adaptation of $q(z)$ (leading to better solution proposals for the current model $p$), and over the model structure $p$ (leading to a better problem representation). 

As an aside, an interesting consequence of the FE decomposition into problem plus solution costs is that a relatively poor problem representation with a superior inference process may be preferred (evidenced by lower FE), over a model with a good problem representation (high Bayesian evidence) where inference costs are high. The notion that the model with the largest Bayesian evidence may not be the most useful model in a practical application, casts an interesting light on the common interpretation of FE as a mere upper bound on Bayesian evidence. We argue here that FE is actually a more principled performance score for a model, since in addition to Bayesian model evidence, FE also scores the performance loss in a model due to an inaccurate inference process. 

\subsection{AIF for smart data sets and resource management}

If we want the robot to cope with unknown physical terrain conditions, it is not sufficient to pre-train the robot offline on a large set of relevant examples. The robot must be able to acquire relevant new data and update its model under real-world conditions.  

FE minimization in the generative model's roll-out to the future results in the minimization of a cost functional known as the Expected Free Energy (EFE). It can be shown that the EFE decomposes into a sum of pragmatic (goal-driven, exploitation) and epistemic (information-seeking, exploration) costs \cite{friston_active_2015}. As a result, inferred actions balance the need to acquire informative data (to learn a better predictive model) with the goal to reach desired future behavior. 

In contrast to the current AI direction towards training larger models on larger data sets, an active inference process elicits an optimally informative, small (``smart'') data set for training of just ``good-enough'' models to achieve a desired behavior. AIF agents adapt enough to accomplish the task at hand while minimizing the consumption of resources such as energy, data, and time. The trade-off between data accuracy and resource consumption is driven by the decomposition in \eqref{eq:complexity-accuracy} of FE as a measure of complexity minus accuracy. According to this decomposition, more accurate models are only pursued if the increase in accuracy outweighs the resource consumption costs.   

In short, AIF agents that are driven solely by FE minimization will inherently manage their computational resources. These agents automatically infer actions that elicit appropriately informative data to upgrade their skills toward good-enough performance levels. Since both the agent and environment mutually affect each other in a real-time information processing loop, it would not be possible to acquire the same data set through the sampling of the environment without the agent's participation.


\section{FE Minimization by Reactive Message Passing}\label{sec:RMP}

\subsection{Why message passing-based inference?}\label{sec:why-MP}

Up to this point, our arguments strongly supported AIF as an information processing engine for the robot. Unfortunately, the computational demands for simulating a non-trivial synthetic AIF agent are extreme. For comparison, consider the human brain that minimizes in real-time, for less than 20 watts, a highly time-varying FE functional (visual data rate about of about a million bits per second) over about $100$ trillion latent variables (synapses). It has been estimated that the human brain consumes about a million times less energy than a high-tech silicon computer on quantitatively comparable information processing tasks. \cite{smirnova_organoid_2023}. 

Clearly, the human brain minimizes FE in a very different way than is available in standard optimization toolboxes. In this section, we will argue for developing a FE minimization toolbox based on reactive message passing in a factor graph.   

\begin{figure}[tb]
\centering
\includegraphics[width=0.95\columnwidth]{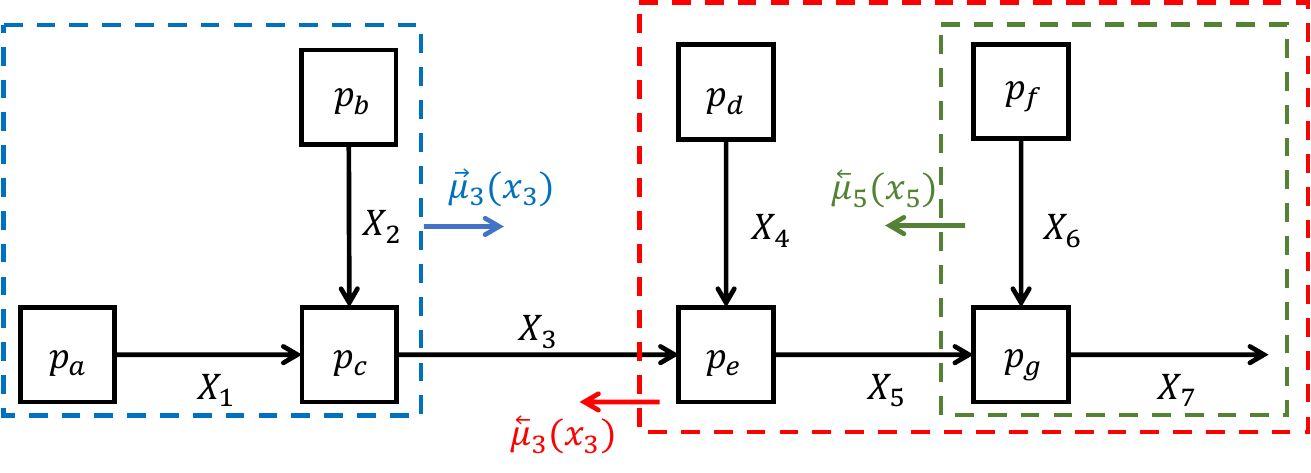}
\caption{Forney-style Factor Graph representation of the factorization \eqref{eq:factorized-model}.}
\label{fig:example-ffg}
\end{figure}

First, we shortly recapitulate why message passing in factor graphs is an effective inference method for large models. Consider a factorized multivariate function
\begin{align}\label{eq:factorized-model}
p(x_1,&x_2,\ldots,x_7) \notag \\
&= f_a(x_1) f_b(x_2) f_c(x_1,x_2,x_3) f_d(x_4) f_e(x_3,x_4,x_5) f_f(x_6) f_g(x_5,x_6,x_7)
\end{align}
Assume that we are interested in inferring (the so-called marginal distribution) 
\begin{equation}\label{eq:x3-marginal}
  p(x_3) = \sum_{x_1}\sum_{x_2}\sum_{x_4}\sum_{x_5}\sum_{x_6}\sum_{x_7} p(x_1,x_2,\ldots,x_7)  
\end{equation}
If each variable $x_i$ in \eqref{eq:x3-marginal} has about $10$ possible values, then the sum contains about $1$ million terms. However, making use of the factorization \eqref{eq:factorized-model} and the distributive law \cite{noauthor_distributive_2022}, we can rewrite this sum as 
\begin{align}\label{eq:x3-marginal-by-mp}
p(&x_3) = 
\bigg( \overbrace{\sum_{x_1}\sum_{x_2} f_a(x_1) f_b(x_2) f_c(x_1,x_2,x_3)}^{\overrightarrow{\mu}_3(x_3)} \bigg) \cdot \notag \\
&\cdot \bigg( \underbrace{\sum_{x_4} \sum_{x_5} f_d(x_4) f_e(x_3,x_4,x_5) \big( \overbrace{\sum_{x_6} \sum_{x_7} f_f(x_6) f_g(x_5,x_6,x_7)}^{\overleftarrow{\mu}_5(x_5)}\big)}_{\overleftarrow{\mu}_3(x_3)}\bigg)
\end{align}

The computation in \eqref{eq:x3-marginal-by-mp}, which requires only a few hundred summations and multiplications, is clearly preferred from a computational load viewpoint. To execute \eqref{eq:x3-marginal-by-mp}, we need to compute intermediate results $\overrightarrow{\mu}_{i}(x_i)$ and $\overleftarrow{\mu}_{i}(x_i)$ that afford an interpretation of local messages in a Forney-style Factor Graph (FFG) representation of the model, see Fig.~\ref{fig:example-ffg}.

Variational FE minimization can also be executed by message passing in a factor graph. In fact, nearly all known effective variational inference methods on factorized models can be interpreted as minimization of a so-called ``constrained Bethe Free Energy'' (CBFE) functional \cite{senoz_variational_2021}. In this formulation, posterior variational beliefs are factorized into beliefs over both the nodes and the edges of the graph. It is possible to add constraints to these local beliefs such as requiring that a particular variational posterior is expressed by a Gaussian distribution. In general, CBFE minimization by message passing in a factor graph supports local adaptation of a plethora of constraints to optimize accuracy vs resource consumption. \cite{senoz_variational_2021, akbayrak_extended_2021}   

Useful dynamic models for real-time processing of data streams with a large number of latent variables are necessarily sparsely connected because otherwise, real-time inference would not be tractable. In sparse models, the computational complexity of inference can be vastly reduced by message passing in a factor graph representation of the model. In particular, automated CBFE minimization by message passing in a factor graph supports refined optimization of the accuracy vs resource consumption balance.

\subsection{Reactive vs procedural coding style}\label{sec:Reactive-vs-Procedural}

Next, we discuss a key technological component for a synthetic AIF agent, namely the requirement to execute FE minimization through a \emph{reactive} programming paradigm.

A crucial feature of all MP-based inference is that the inference process consists entirely of a (parallelizable) series of small steps (messages) that individually and independently contribute to FE minimization. As a result, a message passing-based FE minimization process can be interrupted \emph{at any time} without loss of important intermediate computational results.    

In a practical setting, it is very important that an ongoing inference process can be robustly (without crashing) interrupted at any time with a result. These intermediate inference results can only be reliably retrieved if the inference process iteratively updates its beliefs in small steps, or, in other words, by message passing. Moreover, the inference process should not be subject to a prescribed control flow that contains for-loops. Rather, if we were to write code for an anytime-interruptable inference process in a programming language, we should use a \emph{reactive} rather than the more common \emph{procedural} programming style. In a reactively coded inference engine, there is no code for control flow, such as ``do first this, then that'', but instead only a description of how a processing module (a factor graph node) should react to changes in incoming messages. We will call this process \emph{Reactive Message Passing} (RMP) \cite{bagaevReactiveMessagePassing2023}.  In an RMP inference process, there is no prescribed schedule for passing messages such as the Viterbi or Bellman algorithm. Rather, an RMP inference process just \emph{reacts} by FE minimization whenever FE increases due to new observations. 

In Fig.~\ref{fig:AIF-algorithms}, we display the consequences of choosing a reactive programming style for an application engineer like Sarah. The procedural programming style in Algorithm-1 requires Sarah to provide the control flow (the ``procedure'') for the inference process. Sarah needs to write code for when to collect observations, when to update states, etc. The specific control flow in Algorithm-1 is just an example and there exists literature that aims to improve the efficiency of the control flow \cite{champion_realizing_2021, friston_sophisticated_2021}. In order to write an efficient inference control flow recipe for a complex AIF agent, Sarah needs to be an absolute expert in this field. 

Consider in contrast the code for reactive inference in Algorithm-2. In a reactive programming paradigm, there is no control flow. Rather, the only inference instruction is for the agent to react to any opportunity to minimize FE. When FE minimization is executed by a reactive message passing toolbox, the application engineer only needs to specify the model.

Aside from lowering the competence bar for application engineers to design effective AIF agents, the procedural style of implementing FE minimization is fundamentally inappropriate. The control flow in Algorithm-1 necessarily contains many design choices that only become known during deployment. For instance, how far should the agent roll out its model to the future for computing the EFE? This kind of information is highly contextual and not available to the application engineer. In contrast, the application engineer's code for reactive inference ("react to any FEM opportunity") works for any model in any context. In a reactive inference setting, the appropriate planning horizon is going to be continually updated (inferred) with contextual information. In other words, it is the reactive FEM process itself that leads to optimizing the inference control flow.

\begin{figure}[!tb]
\begin{minipage}[t]{0.46\textwidth}
\begin{algorithm}[H]
    \caption{Procedural AIF}\label{alg:procedural}
    \begin{algorithmic}[1]
      \State \textbf{Specify model}
      $p(x,s,u,\theta)$ 
      \For{$t=1,2,\ldots$ } \Comment{Deploy} 
        \State \text{Collect new observation} $x_t$
        \State \text{Update state} $q(s_t|x_{1:t})$
        \State \text{Update desired future} $\tilde{p}(x_{>t})$  
        \State \text{Upd. candidate policies} $\{\pi^{(i)}\}$ 
        \ForAll{$\pi^{(i)}$}
          \State \text{Predict future} $p(x_{>t}|s_t,\pi^{(i)})$
          \State \text{Compute EFE} $G(\pi^{(i)})$
        \EndFor
        \State \text{Select}  $\pi^* = \underset{\pi \in \{\pi^{(i)}\}}{\arg\min~} G(\pi)$
      \EndFor  
    \end{algorithmic}
\end{algorithm}
\end{minipage}
\hfill
\begin{minipage}[t]{0.46\textwidth}
\begin{algorithm}[H]
    \caption{Reactive AIF}\label{alg:reactive}
    \begin{algorithmic}[1]
      \State \textbf{Specify model} $p(x,s,u,\theta)$
      \While {\textrm{true}} \Comment{Deploy} 
        \State \text{React to any FEM opportunity}
      \EndWhile
    \end{algorithmic}
\end{algorithm}
\end{minipage}
\caption{Pseudo-code for procedural and reactive coding styles for AIF agents.}
\label{fig:AIF-algorithms}
\end{figure}

\subsection{RMP for robustness}\label{sec:robustness}

Since an AIF agent executes under situated conditions, it must perform the FE minimization process robustly in real-time.  Consider an agent whose computational resources are represented by a graph and FE minimization results from executing MP-based inference on that graph. Any MP schedule that visits the nodes in the graph in a prescribed fixed order (as would be the case in a procedural approach to FE minimization) is vulnerable to malfunction in any of the nodes in the schedule. In principle, the FE minimization process needs to stop after such a malfunction and proceed to compute a new MP schedule. Since FE minimization is the \emph{only} ongoing computational process, the robot basically moves blindfolded after a reset. Clearly, for robustness, we need a system that continues to minimize FE, even after parts of the graph break down over time. In a reactive inference framework, collapse of a component is simply a switch to an alternative model structure. The new model may perform better or worse at FE minimization, but there is no reason to stop processing. 

\subsection{RMP for real-time, situated processing}\label{sec:real-time-processing}

An ongoing RMP process can always be interrupted when computational resources have run out on a given platform. In this way, by trading computational complexity (i.e., the number of messages) for accuracy, any RMP-based inference process can be scaled down to a real-time processing procedure, where of course a prediction accuracy price may have to be paid, depending on the available computational resources. In short, FE minimization in any model can be executed in real-time on any computational platform if we implement inference by RMP in a factor graph. 



\subsection{RMP for low power consumption}\label{sec:low-power-consumption}

Similarly, an ongoing RMP process can always be terminated if the expected improvement in accuracy does not outweigh the expected computational load that additional messages would incur.\footnote{The computational load and complexity can only be equated in the absence of a Von Neumann bottleneck (i.e., with mortal computation or in-memory processing). This is because energy and time are ‘wasted’ by reading and writing to memory. \label{fnlabel}} Note that, since FE decomposes as computational complexity minus accuracy, interrupting an RMP-based inference process for this reason is fully consistent with the goal of FE minimization.

Interrupting an ongoing MP process by any of the above-mentioned reasons (e.g., node malfunction, running out of computational resources, expected processing costs outweighing expected accuracy gains, etc.), in principle always leads to sacrificing some prediction accuracy in favor of saving computational costs. Crucially, these interrupts will not cause a system-wide crash in a reactive system.

\section{Model Structure Adaptation}\label{sec:structural-adaptation}

In section~\ref{sec:one-solutiona-approach}, we touched upon the notion that FE minimization should ideally drive the generative model $p$ to evolve to structurally segregated but communicating sub-models that reflect the causal structure of the environment. Technically, this is due to the drive for a lower surprise ($-\log p(x)$).

There is another reason why online structural adaptation is important. Free energy minimization over the structure of $p$ should also lead to a model structure for which inference costs $D_{\text{KL}}[q(z)|| p(z|x)]$ are lower by moving $p(z|x)$ closer to $q(z)$. Consider again the procedural and reactive inference code in Fig.~\ref{fig:AIF-algorithms}. The control flow in the procedural code aims to cleverly steer the inference process toward maximal inference accuracy for minimal computational costs. In contrast, the reactive code just declares that the system should react (by message passing) to any FE minimization opportunity. In the reactive framework, \emph{clever} inference is learned over time by continual minimization over all movable parts of the CBFE, i.e., by FEM over states, parameters, structure (adaptation of $p$), and constraints (adaptation of the structure of $q$). To learn the most effective paths for inference, the toolbox should support structural adaptation over both $p$ and $q$. 


Unfortunately, online structural adaptation during the deployment of the robot is still an ongoing research issue, e.g., \cite{friston_bayesian_2018, loeliger_sparsity_2016, beckers_principled_2022}. One technical difficulty is that an efficient inference control flow (which states are updated at what time, etc.) may change if the structure of the generative model changes. In a procedural programming style, we would need to reset the system and reprogram the inference code in Algorithm-1 (in Fig.~\ref{fig:AIF-algorithms}). This is incompatible with the demand that the agent adapts during deployment. As discussed above, a reactive programming style solves this issue since the application inference code (Algorithm-2 in Fig.~\ref{fig:AIF-algorithms}) is independent of the model structure.

\section{Discussion}\label{sec:discussion}

\subsection{Review of arguments}

We shortly summarize our view on a professional-level supporting software toolbox for the design of relevant AIF agents, see also Table~\ref{tab:technology}. In section~\ref{sec:FEP-and-AIF}, we discussed a few extraordinary features that follow straightaway from committing to free energy minimization as the sole computational mechanism for a future AI ecosystem as proposed in \textcite{friston_designing_2022}. First, the FE functional in an AIF agent can be interpreted as a universal performance criterion that applies in principle to all problems. If FEM can be extended to structural model adaptation, then an AIF agent is naturally able to create and solve sub-problems. Moreover, by virtue of the decomposition of EFE into a sum of information- and goal-seeking costs, AIF agents naturally seek out small "smart" data sets.

In terms of FEM implementation, we asserted that useful models are highly factorized and sparse. Efficient inference in factorized models can always be described as message passing in a factor graph. In particular, nearly all known variants of   highly efficient message passing algorithms for FEM can be formulated in a single framework as minimizing a Constrained Bethe Free Energy (CBFE).

\begin{table}[!b]
\centering
\begin{tabular}{ || c | c | c  || }
\hline 
\hspace{1pt} &
 \large{\textbf{realization technology}} & \large{\textbf{benefits}}  \\ 
 \hline   \hline 
   1  & 
    FEP, AIF
  & one solution approach; \\ & & smart data   \\
 \hline  
2  &  reactive message passing  & low power; \\ & & robustness; \\ & & real-time  \\
\hline 
3  & structural adaptation & problem refinement; \\ & & clever inference   \\   
 \hline 
\end{tabular}
    \caption{Summary of benefits for supporting reactive message passing and structural adaptation in an AIF agent.}
    \label{tab:technology}
\end{table}

We then claimed that a \emph{reactive} rather than procedural processing strategy is essential. Reactive message passing-based (RMP) inference is always interruptible with an inference result, thus supporting guaranteed real-time processing, which is a hard requirement for AIF agents in the real world. In comparison to the more common procedural programming approach to FEM, reactive processing also improves robustness, resource consumption, and the capability to make structural changes without the need for resetting the inference process. 

This latter feature, support for online structural adaptation is also a vital feature of a high-quality AIF toolbox. Online structural adaptation leads to both continual problem representation refinement (by lowering surprise) and to a more efficient inference process.  

\subsection{Review of existing tools}
Currently, there exists a small but vibrant research community on the development of open-source tools for simulating synthetic AIF agents. In this community, a few supporting packages have been released, including SPM \parencite{friston_et_al._spm12_2014}, PyMDP \parencite{heins_pymdp_2022} and \texttt{ForneyLab} \parencite{cox_factor_2019}. The SPM toolbox was originally written by Karl Friston and colleagues, and has developed into a very large set of tools and demonstrations for experimental validation of the scientific output of the UCL team and collaborators. PyMDP is a more recent Python package for simulating discrete-state POMDP models by Conor Heins, Alexander
Tschantz and a team of collaborators. \texttt{ForneyLab.jl} is a Julia package from BIASlab (\url{http://biaslab.org}) for simulating FE minimization by message passing in Forney-style factor graphs. Unfortunately, none of the above-mentioned tools support \emph{reactive} message passing-based inference. Therefore, we believe that these tools will serve the community well as AIF prototyping and validation tools, but they will not scale to support real-time, robust simulation of AIF agents with commercializable value. 

\subsection{Reactive message passing with \texttt{RxInfer}}\label{sec:rxinfer}

More recently, BIASlab has released the open-source Julia package \texttt{RxInfer} (\url{http://rxinfer.ml}) to support an engineer at Sarah's level to develop commercially relevant AIF agents that minimize FE by automated reactive message passing in a factor graph \cite{bagaevReactiveMessagePassing2023}. Julia is a modern open-source scientific programming language with roughly the syntax of MATLAB and out-of-the-box speed of C \cite{bezanson_julia_2017}.

The development process of \texttt{RxInfer} focuses on the following priorities:
\begin{enumerate}
    \item model space coverage
    \begin{itemize}
        \item \texttt{RxInfer} aims to support reactive message passing-based FEM for a very large set of freely definable relevant probabilistic models. 
    \end{itemize}
    \item user experience
    \begin{itemize}
        \item \texttt{RxInfer} aims to support a busy, competent researcher or developer who understands probabilistic modeling (but doesn't know Julia) to design and deploy an AIF agent into the world. In particular, a user-friendly specification of nested AIF agents should be supported. 
    \end{itemize}
    \item adaptation
    \begin{itemize}
        \item \texttt{RxInfer} aims to support continual adaptation by automated FEM over all movable parts of the CBFE functional, including states, parameters, structure, and variational constraints.
    \end{itemize}
    \item real-time
    \begin{itemize}
        \item \texttt{RxInfer} aims to process data streams in ``hard'' real-time, under situated conditions, even for large models. Larger models may lead to less accurate inference (in terms of KL-divergence between variational and Bayesian posteriors), but no crashes.  
    \end{itemize}
    \item low-power
    \begin{itemize}
        \item \texttt{RxInfer} aims to process data streams on any, possibly time-varying, power budget. Lower power budgets may lead to less accurate inference but no crashes.
    \end{itemize}
\end{enumerate}

At the time of writing this paper, \texttt{RxInfer} supports fast and robust automated CBFE minimization by reactive message passing for states and parameters in a large set of freely definable models. \texttt{RxInfer} processes streaming data very fast, but not yet guaranteed in hard real-time. User-friendly specifications of AIF agents will be released this summer. Model structure adaptation is supported by NUV priors (normal priors with unknown variance) \cite{loeliger_sparsity_2016}, but not yet by online Bayesian model reduction \cite{beckers_principled_2022, friston_bayesian_2018}. \texttt{RxInfer} comes with a large set of examples and is slated to support the above priority list in the future.

\section{Conclusions}
Supported by \texttt{RxInfer} or a similar toolbox, future AI engineers will no longer design end-product algorithms, but will instead design the designers (AIF agents) of production algorithms in short and easy-readable code scripts. Along with \cite{friston_designing_2022}, we think that the potential benefits of shared intelligence in ecosystems of  communicating AIF agents are hard to overstate. As we have argued in this position paper, the required underlying technology for realizing this vision is very demanding and currently not yet available. Still, we also think it is not out of reach and is one of the most exciting ongoing research threads in the AI field. 

\subsubsection{Acknowledgments}

I would like to acknowledge my colleagues at BIASlab (\url{http://biaslab.org}) for the stimulating work environment and the anonymous reviewers for excellent feedback on the draft version. Some wording in this document, such as footnote 2, comes straight from a reviewer.

\printbibliography

@inproceedings{loeliger_sparsity_2016,
	location = {La Jolla, {CA}, {USA}},
	title = {On sparsity by {NUV}-{EM}, Gaussian message passing, and Kalman smoothing},
	isbn = {978-1-5090-2529-9},
	url = {http://ieeexplore.ieee.org/document/7888168/},
	doi = {10.1109/ITA.2016.7888168},
	eventtitle = {2016 Information Theory and Applications ({ITA})},
	pages = {1--10},
	booktitle = {2016 Information Theory and Applications Workshop ({ITA})},
	publisher = {{IEEE}},
	author = {Loeliger, Hans-Andrea and Bruderer, Lukas and Malmberg, Hampus and Wadehn, Federico and Zalmai, Nour},
	urldate = {2021-07-21},
	date = {2016-01},
	langid = {english},
}

@article{friston_active_2015,
	title = {Active inference and epistemic value},
	volume = {0},
	issn = {1758-8928},
	url = {http://dx.doi.org/10.1080/17588928.2015.1020053},
	doi = {10.1080/17588928.2015.1020053},
	pages = {null},
	issue = {ja},
	journaltitle = {Cognitive Neuroscience},
	author = {Friston, Karl and Rigoli, Francesco and Ognibene, Dimitri and Mathys, Christoph and {FitzGerald}, Thomas and Pezzulo, Giovanni},
	urldate = {2015-02-22},
	date = {2015-02-17},
	pmid = {25689102},
}

@book{friston_et_al._spm12_2014,
	title = {{SPM}12 toolbox, http://www.fil.ion.ucl.ac.uk/spm/software/},
	author = {Friston et al., Karl J.},
	date = {2014},
}

@article{friston_bayesian_2018,
	title = {Bayesian model reduction},
	url = {http://arxiv.org/abs/1805.07092},
	journaltitle = {{arXiv}:1805.07092 [stat]},
	author = {Friston, Karl and Parr, Thomas and Zeidman, Peter},
	urldate = {2018-05-28},
	date = {2018-05-18},
	eprinttype = {arxiv},
	eprint = {1805.07092},
}

@article{cox_factor_2019,
	title = {A factor graph approach to automated design of Bayesian signal processing algorithms},
	volume = {104},
	issn = {0888-613X},
	url = {http://www.sciencedirect.com/science/article/pii/S0888613X18304298},
	doi = {10.1016/j.ijar.2018.11.002},
	pages = {185--204},
	journaltitle = {International Journal of Approximate Reasoning},
	shortjournal = {International Journal of Approximate Reasoning},
	author = {Cox, Marco and van de Laar, Thijs and de Vries, Bert},
	urldate = {2018-11-16},
	date = {2019-01-01},
}

@article{bezanson_julia_2017,
	title = {Julia: A Fresh Approach to Numerical Computing},
	volume = {59},
	issn = {0036-1445},
	url = {https://epubs.siam.org/doi/10.1137/141000671},
	doi = {10.1137/141000671},
	shorttitle = {Julia},
	pages = {65--98},
	number = {1},
	journaltitle = {{SIAM} Review},
	shortjournal = {{SIAM} Rev.},
	author = {Bezanson, Jeff and Edelman, Alan and Karpinski, Stefan and Shah, Viral B.},
	urldate = {2022-02-03},
	date = {2017-01-01},
	note = {Publisher: Society for Industrial and Applied Mathematics},
}

@article{heins_pymdp_2022,
	title = {pymdp: A Python library for active inference in discrete state spaces},
	url = {http://arxiv.org/abs/2201.03904},
	shorttitle = {pymdp},
	journaltitle = {{arXiv}:2201.03904 [cs, q-bio]},
	author = {Heins, Conor and Millidge, Beren and Demekas, Daphne and Klein, Brennan and Friston, Karl and Couzin, Iain and Tschantz, Alexander},
	urldate = {2022-02-03},
	date = {2022-01-11},
	eprinttype = {arxiv},
	eprint = {2201.03904},
}

@article{senoz_variational_2021,
	title = {Variational Message Passing and Local Constraint Manipulation in Factor Graphs},
	volume = {23},
	issn = {1099-4300},
	doi = {10.3390/e23070807},
	pages = {807},
	number = {7},
	journaltitle = {Entropy (Basel, Switzerland)},
	shortjournal = {Entropy (Basel)},
	author = {Şenöz, İsmail and van de Laar, Thijs and Bagaev, Dmitry and de Vries, Bert},
	date = {2021-06-24},
	pmid = {34202913},
	pmcid = {PMC8303273},
}

@article{bagaevReactiveMessagePassing2023,
	title = {Reactive Message Passing for Scalable Bayesian Inference},
	volume = {2023},
	issn = {1058-9244},
	url = {https://www.hindawi.com/journals/sp/2023/6601690/},
	doi = {10.1155/2023/6601690},
	pages = {e6601690},
	journaltitle = {Scientific Programming},
	author = {Bagaev, Dmitry and de Vries, Bert},
	urldate = {2023-05-28},
	date = {2023-05-27},
	langid = {english},
	note = {Publisher: Hindawi},
}

@book{lanczos_variational_1986,
	location = {New York},
	edition = {4th Revised ed. edition},
	title = {The Variational Principles of Mechanics},
	isbn = {978-0-486-65067-8},
	pagetotal = {464},
	publisher = {Dover Publications},
	author = {Lanczos, Cornelius},
	date = {1986-03-01},
}

@article{friston_sophisticated_2021,
	title = {Sophisticated Inference},
	volume = {33},
	issn = {0899-7667},
	url = {https://doi.org/10.1162/neco_a_01351},
	doi = {10.1162/neco_a_01351},
	pages = {713--763},
	number = {3},
	journaltitle = {Neural Computation},
	shortjournal = {Neural Computation},
	author = {Friston, Karl and Da Costa, Lancelot and Hafner, Danijar and Hesp, Casper and Parr, Thomas},
	urldate = {2022-02-14},
	date = {2021-03-01},
}

@misc{friston_designing_2022,
	title = {Designing Ecosystems of Intelligence from First Principles},
	url = {http://arxiv.org/abs/2212.01354},
	doi = {10.48550/arXiv.2212.01354},
	number = {{arXiv}:2212.01354},
	publisher = {{arXiv}},
	author = {Friston, Karl J. and Ramstead, Maxwell J. D. and Kiefer, Alex B. and Tschantz, Alexander and Buckley, Christopher L. and Albarracin, Mahault and Pitliya, Riddhi J. and Heins, Conor and Klein, Brennan and Millidge, Beren and Sakthivadivel, Dalton A. R. and Smithe, Toby St Clere and Koudahl, Magnus and Tremblay, Safae Essafi and Petersen, Capm and Fung, Kaiser and Fox, Jason G. and Swanson, Steven and Mapes, Dan and René, Gabriel},
	urldate = {2022-12-08},
	date = {2022-12-02},
	eprinttype = {arxiv},
	eprint = {2212.01354 [nlin]},
}

@article{akbayrak_extended_2021,
	title = {Extended Variational Message Passing for Automated Approximate Bayesian Inference},
	volume = {23},
	rights = {http://creativecommons.org/licenses/by/3.0/},
	issn = {1099-4300},
	url = {https://www.mdpi.com/1099-4300/23/7/815},
	doi = {10.3390/e23070815},
	pages = {815},
	number = {7},
	journaltitle = {Entropy},
	author = {Akbayrak, Semih and Bocharov, Ivan and de Vries, Bert},
	urldate = {2023-05-26},
	date = {2021-07},
	langid = {english},
	note = {Number: 7
Publisher: Multidisciplinary Digital Publishing Institute},
}

@article{champion_realizing_2021,
	title = {Realizing Active Inference in Variational Message Passing: The Outcome-Blind Certainty Seeker},
	volume = {33},
	issn = {0899-7667},
	url = {https://doi.org/10.1162/neco_a_01422},
	doi = {10.1162/neco_a_01422},
	shorttitle = {Realizing Active Inference in Variational Message Passing},
	pages = {2762--2826},
	number = {10},
	journaltitle = {Neural Computation},
	shortjournal = {Neural Computation},
	author = {Champion, Théophile and Grześ, Marek and Bowman, Howard},
	urldate = {2023-05-26},
	date = {2021-09-16},
}

@article{smirnova_organoid_2023,
	title = {Organoid intelligence ({OI}): the new frontier in biocomputing and intelligence-in-a-dish},
	journaltitle = {Frontiers in Science},
	author = {Smirnova, Lena and Caffo, Brian S and Gracias, David H and Huang, Qi and Morales Pantoja, Itzy E and Tang, Bohao and Zack, Donald J and Berlinicke, Cynthia A and Boyd, J Lomax and Harris, Timothy D and {others}},
	date = {2023},
	note = {Publisher: Frontiers},
}

@inreference{noauthor_distributive_2022,
	title = {Distributive property},
	rights = {Creative Commons Attribution-{ShareAlike} License},
	url = {https://en.wikipedia.org/w/index.php?title=Distributive_property&oldid=1124679546},
	booktitle = {Wikipedia},
	urldate = {2023-05-26},
	date = {2022-11-29},
	langid = {english},
	note = {Page Version {ID}: 1124679546},
}

@misc{beckers_principled_2022,
	title = {Principled Pruning of Bayesian Neural Networks through Variational Free Energy Minimization},
	url = {http://arxiv.org/abs/2210.09134},
	doi = {10.48550/arXiv.2210.09134},
	number = {{arXiv}:2210.09134},
	publisher = {{arXiv}},
	author = {Beckers, Jim and van Erp, Bart and Zhao, Ziyue and Kondrashov, Kirill and de Vries, Bert},
	urldate = {2023-05-26},
	date = {2022-10-17},
	eprinttype = {arxiv},
	eprint = {2210.09134 [cs, eess]},
}

\end{document}